\documentclass[runningheads]{llncs}
\usepackage[T1]{fontenc}
\usepackage{graphicx}
\usepackage{color}
\usepackage[colorinlistoftodos]{todonotes}
\usepackage{multirow}
\usepackage{amsmath}
\usepackage{algorithm}
\usepackage[noend]{algpseudocode}
\usepackage{cite}
\usepackage{subfig}

\newcommand{\ie}{i.e.\ }
\newcommand{\eg}{e.g.\ }
\newcommand{\eq}{Eq.\ }

\newcommand{\spara}[1]{\vspace{0.1cm}\noindent{\bf #1}}

\newcommand*{\images}{images}%

\newcommand{\etal}{~\emph{et~al.}}

\begin{document}
\title{TextMatcher: Cross-Attentional Neural Network to Compare Image and Text}
%
%


\author{Valentina Arrigoni \and
Luisa Repele \and
Dario Marino Saccavino}

\authorrunning{Arrigoni et al.}
\titlerunning{Cross-Attentional Neural Network to Compare Image and Text}
%
\institute{
	UniCredit, Italy
	\email{\{valentina.arrigoni,luisa.repele,dariomarino.saccavino\}@unicredit.eu}}
\maketitle              
\begin{abstract}


We study a multimodal-learning problem where, given an image containing a single-line (printed or handwritten) text and a candidate text transcription, the goal is to assess whether the text represented in the image corresponds to the candidate text.
This problem, which we dub \emph{text matching}, is primarily motivated by a real industrial application scenario of automated cheque processing, whose goal is to automatically assess whether the information in a bank cheque (e.g., issue date) match the data that have been entered by the customer while depositing the cheque to an automated teller machine (ATM).
The problem finds more general application in several other scenarios too, e.g., personal-identity-document processing in user-registration procedures.

We devise a machine-learning model specifically designed for the text-matching problem.
The proposed model, termed \emph{TextMatcher}, compares the two 
inputs by applying a novel cross-attention mechanism over the embedding representations of image and text, and it is trained in an end-to-end fashion on the desired distribution of errors to be detected.


We demonstrate the effectiveness of TextMatcher on the automated-cheque-processing use case, where TextMatcher is shown to generalize well to future unseen dates, unlike existing models designed for related problems.
We further assess the performance of TextMatcher on different distributions of errors on the public IAM dataset.
Results attest that, compared to a na\"ive model and existing models for related problems, TextMatcher achieves higher performance on a variety of configurations.

\end{abstract}

\section{Introduction}


The way we interact with the world concerns stimuli from different senses: im\-ages we see, sounds we hear, words we read. 
All these examples correspond to different \emph{modalities} by which information is presented to us. 
The same variability can apply to data presented to a machine, such as images, free text, sounds, videos.
\emph{Multimodal learning} is an active and challenging research area, whose goal is to build machine-learning models capable of processing and exploiting information from multiple modalities~\cite{baltruvsaitis2018multimodal}.
It includes numerous (classes of) tasks -- such as multimodal representation learning, modality translation, multimodal alignment, multimodal fusion, co-learning -- 
and finds application in a wide range of scenarios -- such as audio-visual speech recognition, image/video captioning, media description, multimedia retrieval.


In this paper, we introduce the following multimodal-learning task, which we term \emph{text matching}: 
given an image representing a single line of (printed or handwritten) text and a candidate text transcription, assess whether the text inside the image corresponds to the candidate text.


\spara{Applications.}
The prominent application of the text-matching problem is a real industrial use case of automated cheque processing, which naturally arises in the banking domain.
In this context, a customer of a bank deposits a bank cheque to an automated teller machine (ATM).
While inserting the cheque into the ATM, the customer is typically required to also type (through the ATM keypad) some information that is written on the cheque, such as issue date, amount, and beneficiary.
The match between what is actually written in the cheque and the data entered by the user is a-posteriori verified by back-office operators, who would clearly benefit from a decision support system that has at its core a method to perform this check automatically.

Text matching finds applications in several other real-world scenarios too, in which an image containing text is assigned the (supposedly) corresponding text, and particular kinds of mismatching must be avoided.
As an example, softwares for user-registration procedures typically need to collect information regarding personal identity documents. 
The user is asked to provide an image of her document and also to enter data that are written in the document, such as document identifier, expiration date, and so on.
Again, back-office operators later-on check if there is a match between the document and the entered data, and, based on the outcome of the match, they accept or reject the registration.

\spara{Challenges.}
An immediate yet na\"ive method to solve the text-matching task is to resort to the related well-established problem of \emph{text recognition}, whose goal is, given an image that is assumed to contain text, to recognize and output the text therein~\cite{chen2020text}.
Specifically, the idea would be to use a text-recognition method to extract the text within the input image and then simply compare the extracted text with the candidate text. 
This is a rather simplistic approach, as it disregards the availability of a candidate text at all.
We claim that designing ad-hoc methodologies for text matching, which properly exploit the information of the candidate text and are specifically trained on the desired distribution of non-matching texts, can be more effective. This claim is experimentally confirmed, see Section~\ref{sec:exps} for more details.

\spara{Contributions.}
We tackle the text-matching problem by devising a machine-learning model that is specifically designed for it.
The proposed model, dubbed \emph{TextMatcher}, 
scans the input image horizontally, searching for characters of the candidate text. 
This is performed by projecting the input image and text into separate \emph{embedding spaces}.
Then, a novel \emph{cross-attention mechanism} is employed, which aims to discover local alignments between the characters of the text and the vertical slices of the image. 
The ultimate similarity score produced by the model is a weighted cosine similarity between features of the characters and features of the slices of the image, where the weights are the computed attention scores. Such a score is eventually used to answer the original yes/no matching question via a thresholding approach.

The model is trained in an end-to-end fashion and, thanks to the cross-attention mechanism, it produces consistent embedding spaces for both image and text, and it is able to successfully specialize to specific distributions of errors.
This is desirable because, depending on the application, it can be appropriate to either correct minor typos or enforce the exact spelling of every word. 

\spara{Summary and roadmap.} To summarize, our main contributions are:
\begin{itemize}
\item We study a multimodal-learning problem termed \emph{text matching} (Section~\ref{sec:probdef}), which finds application in a variety of real scenarios, including an industrial use case of automated cheque processing, peculiar of the banking domain.
\item We devise a machine-learning model, termed \emph{TextMatcher}, that is specifically designed for text matching and exploits a novel cross-attention mechanism (Section~\ref{sec:algs}).
\item We showcase the proposed TextMatcher in the primary application context of automated cheque processing, by carrying out experiments on a real-world (proprietary) dataset of bank cheques provided by UniCredit, a noteworthy pan-European commercial bank (Section~\ref{sec:experiments_real}).\footnote{TextMatcher has been deployed at UniCredit, and it is currently used in production.}
\item We further test the performance of TextMatcher on the popular public IAM dataset~\cite{marti2002iam} (Section~\ref{experiments:iam}).
Results on both UniCredit and IAM datasets attest that TextMatcher achieves high accuracy and is capable of properly handling specific distributions of errors.
It also consistently outperforms a na\"ive model and existing text-recognition methods in both those aspects.



\end{itemize}

\noindent
Section~\ref{sec:relatedwork} overviews the related literature. Section~\ref{sec:conc} concludes the paper.

\section{Related Work}


\label{sec:relatedwork}

The problem we tackle in this work, i.e., text matching, falls into the broad area of multimodal learning. A comprehensive survey of the main challenges, problems, and methods in this area is provided by Baltruvsaitis~\emph{et al.}~ \cite{baltruvsaitis2018multimodal}.
Referring to the taxonomy reported in that survey, the category that better complies with text matching is the \emph{(implicit) alignment} one, which encompasses multimodal-learning problems whose goal is to identify relationships between sub-elements from different modalities, possibly as an intermediate step for another task.

To the best of our knowledge, the text-matching problem has not been specifically studied in the literature: no ad-hoc method has been designed for it so far.
Nevertheless, there exist tasks/methods that share some similarities.
In the remainder of this section, we overview such related works.

\spara{Text recognition.}
Recognizing text in images has been an active research topic for decades.
A plethora of different approaches exist.
A prominent state-of-the-art text-recognition model, which we take as a reference in this work, is ASTER~\cite{ShiWLYB16,shi2018aster}, i.e.,  an end-to-end neural network that is based on an attentional sequence-to-sequence model to predict a character sequence directly from the input image. For more approaches and details on text recognition, we refer to comprehensive Chen~\emph{et al.}'s survey~\cite{chen2020text}.

The main difference between text recognition and our text-matching problem is that the former extracts text from images \emph{without relying on any input candidate text}.
A na\"ive approach to text matching would be to run a text-recognition method on the input image, and using the input candidate text only to check the correspondence with the recognized text.
A major limitation of this approach is that it disregards the candidate text at all, thus resulting intuitively less effective than approaches that, like the proposed TextMatcher, are specifically designed for text matching and profitably exploit the candidate text and the given distribution of errors to be recognized.
More specifically, the technical strengths of the proposed TextMatcher method over a text-recognition-based approach are:
\begin{itemize}
	\item While the text-recognition model is trained only on the matching data, TextMatcher is trained with both positive and negative examples, allowing it to better learn the frontier between the two sets whenever it is relevant, for instance when a difference of a single character in a specific position is important for a large portion of data (\eg ``\emph{MR Smith}'' vs. ``\emph{MS Smith}'').
	\item More importantly, we experimented that the TextMatcher model better generalises to different distributions at inference time thanks to the training through negative matching pairs (see Section \ref{sec:experiments_real}). 
	\item If the text-recognition model uses an encoder-decoder architecture (like \cite{shi2018aster}), the corresponding text-matching model needs only the encoder part, therefore it tends to be faster during inference.
\end{itemize}

\spara{Word spotting.}
Given a collection of images representing single words and a query text, word spotting aims at ranking all the images of the word-image collection based on their similarity to the given query~\cite{rath2007word}.
The variant of word spotting where the query is a text string (and not an image), termed \emph{Query-by-String} (QbS), is the one more relevant for our work.
Existing approaches to QbS word spotting aim at learning a map from textual representation to image representation. 
A popular choice for text representation in word spotting is the binary attribute representation referred to as \emph{Pyramidal Histogram of Characters} (PHOC) \cite{almazan2014word}.
Recent works use neural networks in order to learn the mapping from word images to PHOC~\cite{sudholt2016phocnet, sudholt2018attribute, gomez2017lsde, mhiri2019word}. For instance, Sudholt~and~Fink~\cite{sudholt2016phocnet} propose the PHOCNet model, which applies a convolutional neural network (CNN) to word images in order to estimate a probability distribution over attributes of the PHOC representation. 
A more sophisticated approach is proposed by Mhiri\etal~\cite{mhiri2019word}, which learns a mapping from the word images using a CNN and from the input text query with a recurrent neural network (RNN) to a common embedding space, driven by the PHOC representation. On top of the learned embeddings, a matching model is trained to refine the response of the nearest-neighbor queries. Although sharing some similarities with our approach, this method \emph{is not trained end-to-end} together with the matching model, and compares image and text \emph{only after the embedding vectors are produced}, differently from our cross-attention mechanism. 
More importantly, in general, the word-spotting task is usually designed and evaluated with the assumption that the vocabulary of words \emph{is fixed}, which is not true in our case.


\spara{Image-text matching}
is another (loosely) related task, whose goal is to measure the semantic similarity between an image and a text~\cite{stackedcrossattention,selfattention,radford2021learning,PFASforITM}.
Despite similar in spirit, image-text matching is different from text matching from a conceptual point of view.
The fundamental difference is that the input images to image-text matching are \emph{general-purpose} ones, i.e., they are not constrained to represent a (single-line) text.
For instance, the goal in image-text matching might be to assess whether an image depicting a dog playing with a ball is well described by the ``\emph{A dog is playing with a ball}'' text.
For this reason, image-text matching considers the semantic content of the image, whereas text matching looks solely at (the syntax of) the text in the image.
As a result, image-text matching is typically employed in applications far away from the ones targeted by text matching (e.g., generation of text descriptions from images or image search), and existing approaches to image-text matching cannot (be easily adapted to) work for our text-matching problem.
From a methodological point of view, image-text matching and text matching share more similarity, as both the problems can be approached with techniques that involve learning a shared representation for image and text.
However, important technical differences still remain. 
Among the prominent models for image-text matching are the ones proposed in 
\cite{stackedcrossattention,selfattention}, which use a cross-attention mechanism to inspect the alignment between image regions and words in the sentence, 
and \cite{PFASforITM}, which exploits the correlation of semantic roles with positions (those of objects in an image or words in a sentence).
The proposed TextMatcher uses attention as well, 
but, unlike \cite{selfattention,PFASforITM}, it makes a simpler consideration of the horizontal position of a character in the image. Also, while \cite{stackedcrossattention,selfattention,PFASforITM} use pretrained models to generate feature representations for the image regions, our TextMatcher is trained end-to-end, thus being capable of learning the weights of the convolutional layer alongside the attention layer. 
%

\section{Text Matching Problem}
\label{sec:probdef}

\begin{figure}[t]
	\centering
	\vspace{-1mm}
	\includegraphics[width=0.45\textwidth]{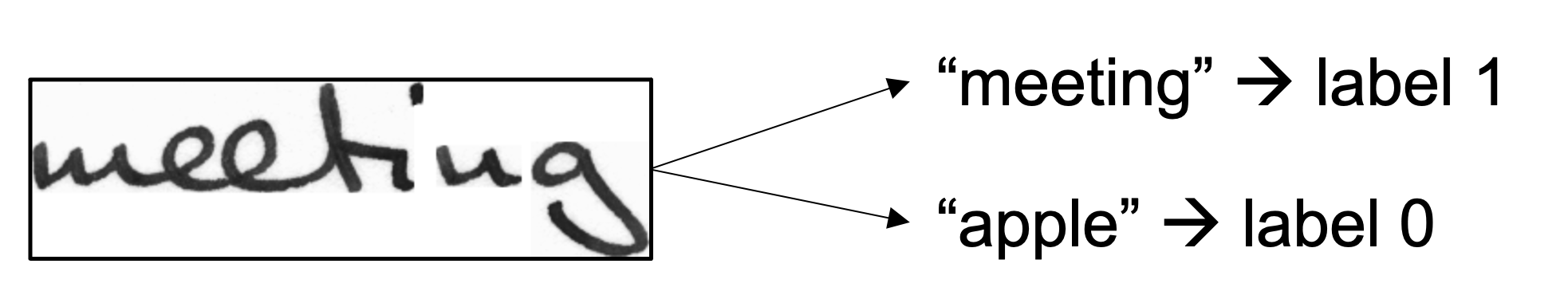}
	\vspace{-4mm}
	\caption{\small Text matching as a binary supervised-classification task.}
	\label{fig:task}
\end{figure}

We tackle a multimodal-learning problem, which we term \emph{text matching} and define as follows: given an image containing a single-line text (printed or handwritten) written horizontally, together with a candidate text transcription, assess whether the text inside the image corresponds to the candidate text. 
This corresponds to a binary supervised-classification task, in which we are given a dataset of the form $\{\left((I^i, t^i),\mbox{ } l^i\right)|i=1, \dots , n \}$, where $I^i$ and $t^i$ are image and text inputs of the $i$-th example, and $l^i$ is the corresponding binary label.
In particular, we adopt the following convention: an (\emph{image}, \emph{text}) pair is assigned the ``$1$'' label if \emph{image} and \emph{text} correspond, and, in this case, the pair is referred to as a \emph{matching pair}. 
Otherwise, the pair is assigned the ``$0$'' label, and it is referred to as a \emph{non-matching pair}.
Similarly, we talk about \emph{matching} and \emph{non-matching} texts for a given image.
An illustration of the input to text matching is in Figure~\ref{fig:task}.

\section{Proposed Approach}
\label{sec:algs}

\begin{figure}[t]
	\centering
	\vspace{-1mm}
	\includegraphics[width=0.8\textwidth]{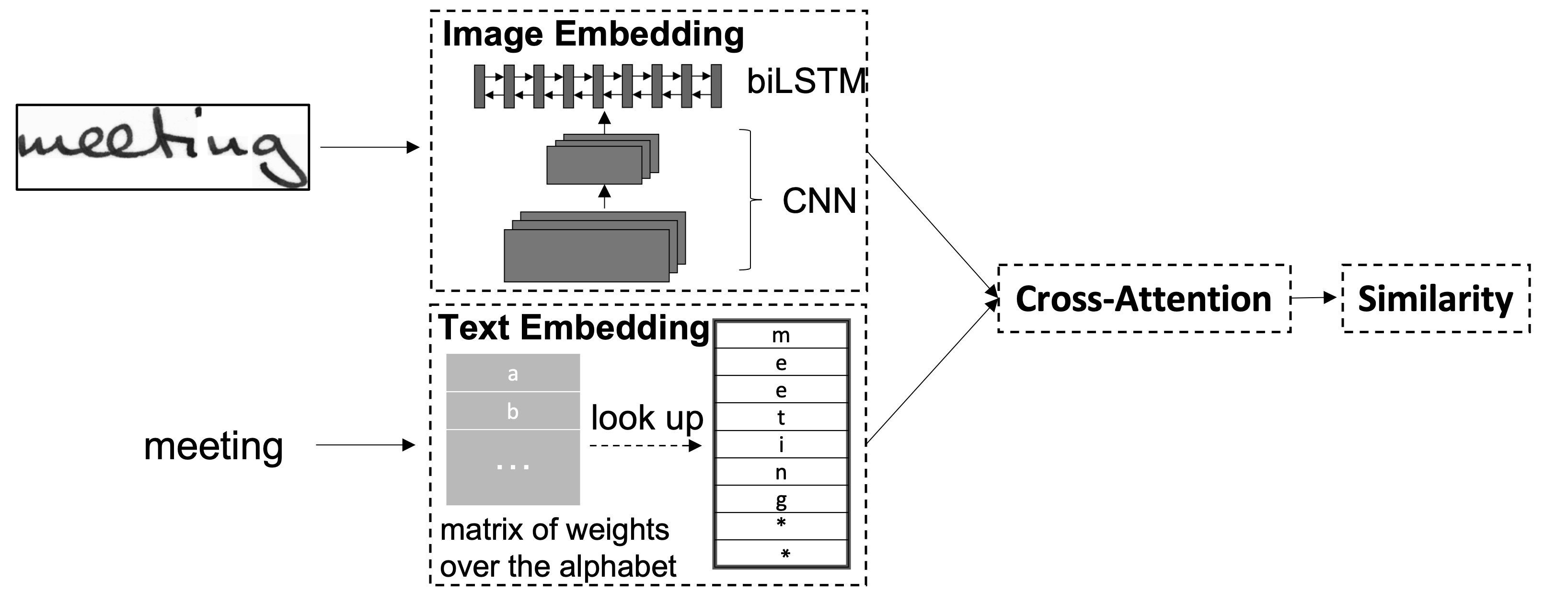}
	\vspace{-4mm}
	\caption{\small TextMatcher architecture.}
	\label{fig:tm_model}
\end{figure}

We propose a model called TextMatcher which directly compares an input image and a candidate text, producing a similarity score. The overall architecture is illustrated in Figure~\ref{fig:tm_model}.
The image and the text are independently projected as matrices into separate embedding spaces, through \emph{image embedding} and \emph{text embedding} blocks, respectively.
These embeddings are then compared with each other through a \emph{cross-attention mechanism}, whose aim is to discover local alignments between the characters of the text and the vertical slices of the image (\ie rectangular regions obtained by scanning the image along the horizontal axis), and produces in output a similarity score.
A key peculiarity of the cross-attention component is that it helps the model specialize to  the distribution of specific errors to be recognized.
The three blocks in the overall TextMatcher model (i.e.,  image embedding, text embedding, and cross-attention mechanism) are jointly trained in an end-to-end fashion, via a contrastive loss function. In the following, we describe in detail the various components of TextMatcher.


\subsection{Image Embedding}\label{sec:image_embedding}

In order to produce the image embedding, the input image is first resized to a fixed dimension, and then it is processed by some convolutional layers, followed by recurrent layers in order to also encode contextual information. The output of this neural network module is the image embedding $J$ of fixed $s_i \times d_{i}$ dimensionality, where $s_i$ denotes the number of receptive fields, or slices, from the input image, and $d_i$ is the feature dimension.
More precisely, in this work we use the encoder block of the ASTER model from \cite{shi2018aster} to extract the image embedding: the input image is fed into a set of convolutional layers and batch normalization layers, followed by a bidirectional Long Short-Term Memory (LSTM) module. All the weights of the convolutional layers, batch normalization layers, and bidirectional LSTM are jointly learned in the final multimodal task.


\subsection{Text Embedding}\label{sec:text_embedding}
As for text embedding, we simply use an embedding matrix over the characters of the alphabet. 
Let $\mathcal{A}$ be the alphabet at hand, which we assume to include a special character for the padding. 
The embedding matrix $T_{emb}$ is a learnable matrix of dimension $|\mathcal{A}|\times d_{t}$. Given a text $c_1 c_2 \dots c_l$, we first pad it to a fixed length $s_t$ (or truncate it, if $l > s_t$). 
Then each character $c_i$ is projected into the embedding space through the embedding matrix $T_{emb}$.
The final embedding of the input text is a matrix $T$ of dimensionality $s_t \times d_t$, whose row $i$ is the row of $T_{emb}$ corresponding to the character $c_i$, for $i=1, \dots, s_t$.

\subsection{Cross-attention Mechanism}\label{sec:attention}

The attention mechanism was originally devised by Bahdanau\etal~\cite{bahdanau2014neural} in the context of encoder-decoder-based machine-translation systems, as a solution to the renowned issue of RNNs of rapid performance degrading as the length of the input sentence increases.
Later, this mechanism has been widely employed in other contexts concerning sequential inputs, including natural language processing, computer vision, and speech processing.

The TextMatcher cross-attention component takes as input the $J$ and $T$ embeddings of image and text, and discovers local alignments between them. 
A similar idea is exploited in the self-attention mechanism of the well-established Transformer architecture \cite{vaswani2017attention}. In the latter, the self-attention computes a weighted representation of each token \emph{attending to} the entire sentence.
Conversely, in our case a multimodal approach is employed: each character of the input text attends to the vertical slices of the image. Moreover, in \cite{vaswani2017attention}, the attention scores are used to compute a weighted sum of the value vectors of each token in the sentence, while, in our case, the attention scores are used to compute a weighted sum of cosine similarities between each character and the slices of the image, since our goal is to compute a similarity between image and text.

Specifically, the cross-attention mechanism of TextMatcher is as follows.
First of all, in order to inject some positional information, we add independent positional embeddings to both image and text embeddings. The positional embeddings have the same dimension of the corresponding text or image embedding, and, as such, they can be summed up. Inspired by \cite{vaswani2017attention}, we use sine and cosine functions of different frequencies, where each dimension of the positional encoding corresponds to a sinusoid.
The rationale is that this function would allow the model to easily learn to attend by relative positions.
From now on, with a little abuse of notation, we will consider $T$ and $J$ as the text and image embeddings with the addition of the positional embeddings.

\begin{figure}[t]
    \centering
    \vspace{-1mm}
\begin{tabular}{cc}
    \hspace{-3mm}\includegraphics[width=0.325\textwidth]{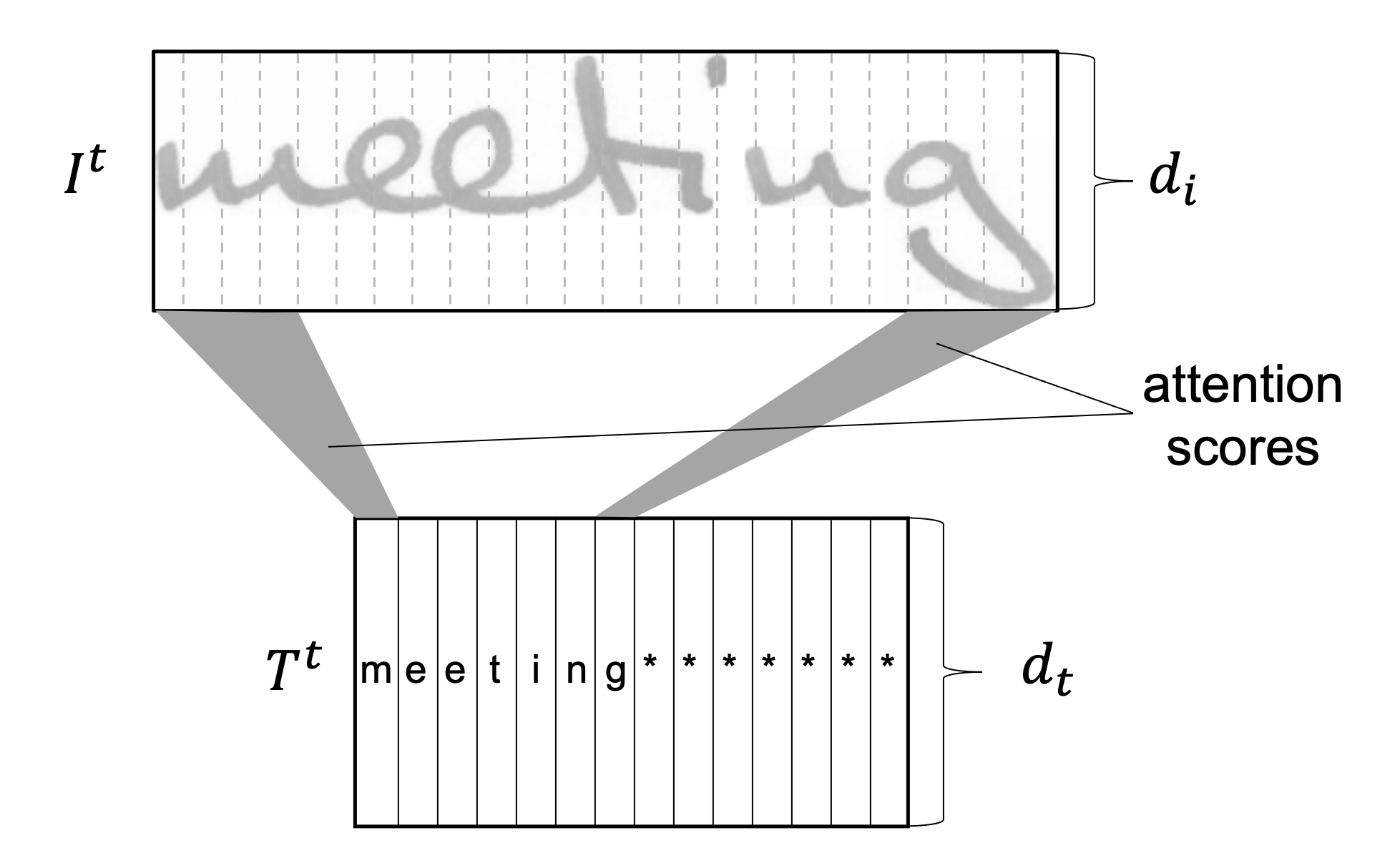} & \qquad \includegraphics[width=0.645\textwidth]{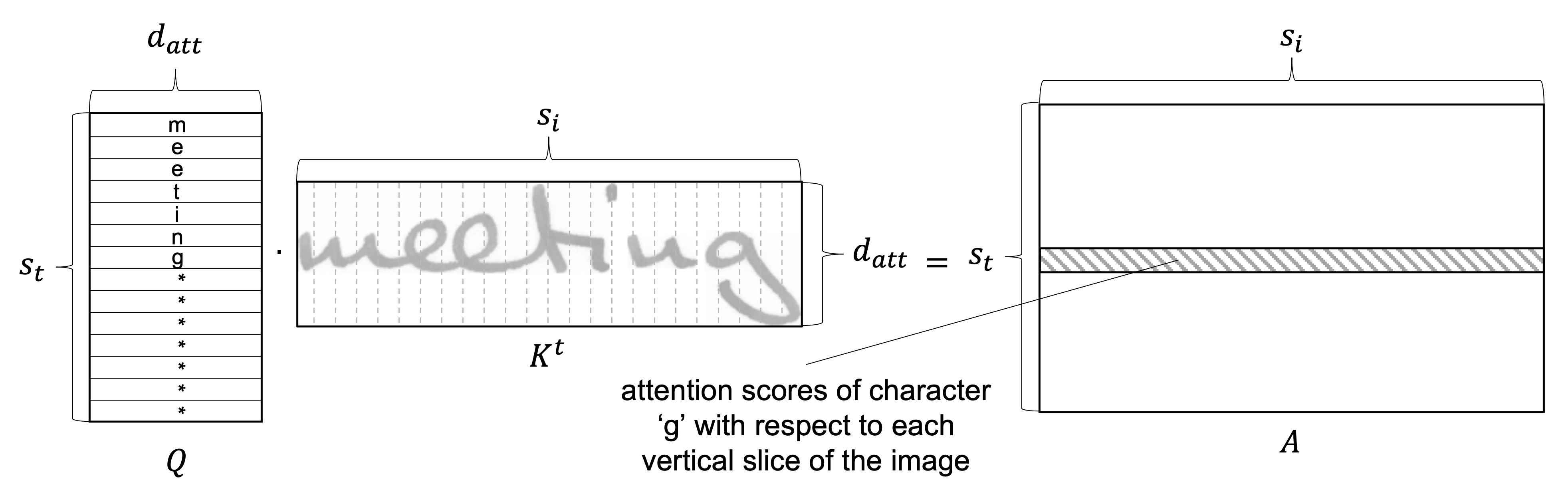}\\
    (a) & \qquad (b)
    \end{tabular}
    \vspace{-2mm}
    \caption{\small(a) Visual representation of the cross-attention mechanism employed in the proposed TextMatcher model. (b) Computation of the attention matrix.\vspace{-4mm}}%
    \label{fig:example}%
\end{figure}


Let us consider the perspective of the text: for each character of the text, we want to compute an attention score with respect to each vertical slice of the image embedding, in order to pay more attention to the portion of the image that is expected to contain the corresponding character.
The idea of this attention mechanism is depicted in Figure~\ref{fig:example}(a). We compute attention scores between the embeddings of the characters and those of the vertical slices of the image by first projecting these vectors into separate embedding spaces of dimension $d_{att}$, and then computing normalized dot products between all pairs of characters and slices of the image. 
In particular, we compute \emph{query} vectors of dimension $d_{att}$ for the text and \emph{key} vectors of dimension $d_{att}$ for the image. These vectors are packed together respectively into the query matrix $Q = T Q_t$ and the key matrix $K= J K_i$, where $Q_t$ and $K_i$ are learnable parameters of dimension $d_{t} \times d_{att}$ and $d_{i} \times d_{att}$ respectively. The resulting matrices are $Q$ of dimension $s_t \times d_{att}$ and $K$ of dimension $s_i \times d_{att}$.
We compute the attention matrix of dimension $s_t \times s_i$ as the dot product between the query $Q$ and the key $K$, and then we apply a softmax function over the columns of the result, as illustrated in Figure~\ref{fig:example}(b):
\begin{equation}
\small
	A=\operatorname{softmax}(Q K^t, \operatorname{dim}=1).
\end{equation}
%
%

\noindent where the notation $\operatorname{dim}=1$ refers to the computation along the columns.
In this way, the $i$-th row of the attention matrix contains the normalized attention scores of the $i$-th character of the input text with respect to each vertical slice of the image embedding.
Then, the value vectors are used to compute a weighted cosine similarity between characters and vertical slices of the image embedding. First, normalized value matrices are computed for both image and text embeddings:
\begin{equation}
\small
	V_{text} = \operatorname{normalize}(T V_t, \operatorname{dim}=1), \qquad V_{image} = \operatorname{normalize}(J V_i, \operatorname{dim}=1),
\end{equation}

\noindent with learnable parameters $V_t$ and $V_i$ of dimension $d_{t} \times d_{att}$ and $d_{i} \times d_{att}$ respectively. The resulting matrices $V_{text}$ of dimension $s_t \times d_{att}$ and $V_{image}$ of dimension $s_i \times d_{att}$ are normalized over the columns in order to directly compute cosine similarities as their dot product. 
The cosine matrix $C=V_{text} V_{image}^t$ has dimension $s_t \times s_i$: the component $(i, j)$ is the cosine similarity between the character at position $i$ and the vertical slice of the image embedding at position $j$. Then, the cosine matrix is multiplied element-wise with the attention matrix, and a sum over the columns is performed, in order to compute a weighted cosine similarity of each character with respect to each vertical slice of the image embedding:
\begin{equation}
\small
	C_{att}=\operatorname{sum}(C \odot A, \operatorname{dim}=1),
\end{equation}
where $\odot$ stands for the element-wise multiplication. Finally, the similarities not related to pad characters are summed up, obtaining the final similarity score between the input image and the candidate text: $S_{tm}=\operatorname{sum}(C_{att}[\operatorname{pad}=1])$. 

%

The $S_{tm}$ score is exploited to ultimately predict the $\hat{l}$ binary label via a thresholding mechanism: given a threshold $\tau$,
\begin{align}\label{eq:threshold}
\small
	\hat{l}=
	\begin{cases}
		1, & \mbox{if }S_{tm} \ge \tau \\
		0, & \mbox{if }S_{tm} < \tau		
	\end{cases}
\end{align}

\subsection{Loss}\label{sec:loss}
Overall,  the TextMatcher model has the following parameters: $W_{encoder}$, $T_{emb}$, $pos_i$, $pos_t$, $Q_t$, $K_i$, $V_t$, $V_i$, where $W_{encoder}$ contains the weights of the image encoder and $pos_t$ and $pos_i$ are the positional embeddings, possibly carefully initialized and then frozen.
Given a dataset of matching and non matching pairs $\{\left((I^i, t^i),\mbox{ } l^i\right)|i=1, \dots , n \}$, where $I^i$ and $t^i$ are image and text inputs of the $i$-th example and $l^i$ is the corresponding binary label, the matching network is trained with the following contrastive loss, originally introduced in \cite{hadsell2006dimensionality} :
\begin{equation}
\small
	L=\alpha l\left(1-S_{tm}\right)^2+\left(1-l \right) \max\{m-\left(1-S_{tm}\right), 0\}^2,
\end{equation}

\noindent where $m$ is the margin and $\alpha$ balances matching and non-matching pairs. 


\section{Experiments}
\label{sec:exps}



In this section, we present experiments to empirically assess the performance of our TextMatcher model, on both an industrial use case of automated cheque processing -- using a real (proprietary) dataset of bank cheques, and on a more general context -- using a popular public dataset of handwritten text (i.e., IAM).



\spara{Competitors.}
We compare our TextMatcher to  ($i$) a text-recognition model adapted to work for text matching, and ($ii$) a na\"ive model for text matching. 

As for the former, the text transcription $\hat{t}$ produced by a text-recognition model run on an input image $I$ is compared to the input candidate text $t$ so as to produce a similarity score.
Specifically, such a similarity score is computed as $S_{tr}=1-\frac{\operatorname{Lev}\left(\hat{t}, t\right)}{\max\{|\hat{t}|, |t|\}}$, where $\operatorname{Lev}\left(\cdot, \cdot \right)$ denotes the well-known \emph{Levenshtein} distance between text strings, and $|t|$ is the number of characters in $t$. The training is performed with the dataset $\{\left(I^i, t^i\right)|i=1, \dots , n \}$, where $I^i$ is the $i$-th image and $t^i$ the corresponding (correct) text.
As a text-recognition model we use the well-established state-of-the-art ASTER~\cite{shi2018aster}.


As a na\"ive text-matching model, we consider a model that computes image and text embeddings separately, and then computes the cosine similarity between their average vectors. The image embedding $J$ of dimension $s_i \times d_i$ and the text embedding $T$ of dimension $s_t \times d_t$ are defined in the same way as in TextMatcher, with the constraint $d_i=d_t$.
Then, the average embeddings $\mathbf{T}_{avg}=\operatorname{mean}(T, \operatorname{dim}=0)$ and $\mathbf{J}_{avg}=\operatorname{mean}(J, \operatorname{dim}=0)$ are computed, with the convention that rows related to pad characters are not considered in the average of the text embedding. Finally, the output of the model is the cosine similarity between the average image and text $S_{n}=\frac{\mathbf{T}_{avg}^t \cdot \mathbf{J}_{avg}}{ \left \|\mathbf{T}_{avg}\right \| \cdot  \left \| \mathbf{J}_{avg}\right \|}$.
The parameters of the convolutional part and the embedding matrix for the text are trained end-to-end in the final multimodal task, using the loss in Section \ref{sec:loss}.
We term such a na\"ive text-matching model \emph{Na\"iveTextMatcher}.

For both competitors, we compute the predicted binary label from the $S_{tr}$ and $S_n$ similarities  
in the same way as done in TextMatcher (\eq \eqref{eq:threshold}).



\spara{Implementation details.}
We resize each grayscale image to $32\times256$ pixels and normalize pixel values to $[-1.0,1.0]$.
The image embedding part is the encoder of ASTER \cite{shi2018aster} with a final bidirectional LSTM with $256$ hidden dimension, which produces an image embedding of dimension $64 \times 512$.
The encoder is initialized with the weights of the pretrained model published together with the source code of \cite{shi2018aster}.
For the text embedding we use $d_t=512$. 
We add positional embeddings to both image and text embeddings, using the same initialization strategy proposed in  \cite{vaswani2017attention}, and then we freeze them during training. The attention dimension $d_{att}$ is set at 512. The text embedding and the other attention parameters are initialized with the Xavier initialization \cite{glorot2010understanding}. The training is performed with Stochastic Gradient Descent (SGD), with $0.9$ momentum, using a learning rate equal to $0.005$ and batch size $8$. As for the the loss, we use margin $m = 1$ and $\alpha=1$ for the experiments on IAM and $\alpha = 0.2$ for the real use case. The maximum number of epochs is $50$. Na\"iveTextMatcher is initialized analogously to TextMatcher.

As for the text-recognition competitor, we use the available source code of ASTER~\cite{shi2018aster}.\footnote{https://github.com/ayumiymk/aster.pytorch} ASTER is initialized with the weights of the publicly available pretrained model. All hyperparameters are set to the default values, except for the batch size set to $64$, the height of input images set to $32$, and the maximum number of epochs set to $35$. 

In the automated-cheque-processing use case we use an object-detection model (specifically, the state-of-the-art YoLo~\cite{redmon2016you}) to first extract the part of a cheque containing the field of interest.


\subsection{Industrial Use Case of Automated Cheque Processing}\label{sec:experiments_real}
A major application of the text-matching problem is in the context of automated cheque processing.
The typical scenario here is that a customer of a bank who deposits a cheque to an ATM is asked to type some information that is written on the cheque, e.g., amount, issue date, beneficiary. 
Later, back-office operators manually check the correspondence between the information typed by the customer and what is written on the cheque.
The main goal of a text-matching solution is to automate such a correspondence verification, in order to help operators  perform their manual checks more easily and faster.

We evaluated the proposed TextMatcher in automated cheque processing by using a (proprietary) real dataset provided by UniCredit, a renowned pan-European commercial bank. 
Although we experimented with other fields too (i.e., amount, beneficiary), here we focus on the verification of the \emph{issue date} of a bank cheque, as it is more challenging and appropriate to showcase the usefulness of text matching.
Specifically, as main challenges, a model for matching the date field must be \emph{sensitive to single-digit differences}, and the set of texts available at training time (built from cheques deposited in the past) is  \emph{disjoint from the set of texts that will be used at inference time}, as the latter includes \emph{dates that are in the future with respect to the dates observed during training}. 
In order to handle the future-date issue, we take particular advantage of the proposed text-matching framework, which offers the chance to \emph{inject into the model the desired behaviour through the non-matching texts}.
Moreover, the number of non-matching (i.e., negative) samples in the real dataset is very small, therefore we keep these examples for testing purposes and use a synthetic matching dataset for training.
In the following, we first describe the strategy for generating synthetic negative samples, and, then, we present the results of the evaluation.

\begin{figure}[t]
\vspace{-2mm}
	\begin{center}
		\includegraphics[width=0.8\textwidth]{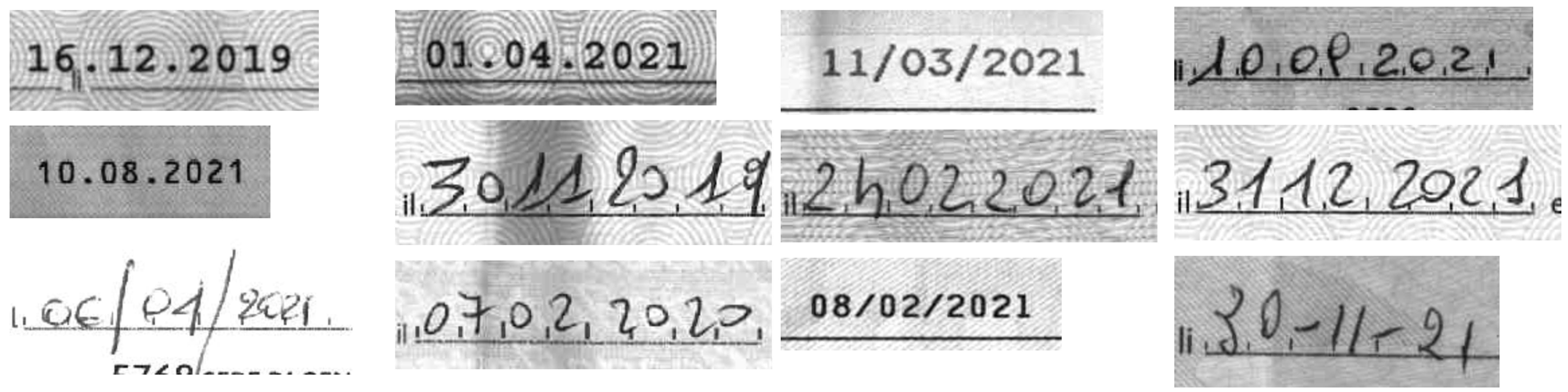}
	\end{center}
	\vspace{-4mm}
	\caption{\small Examples of images representing issue dates of bank cheques.\vspace{-4mm}}
	\label{fig:sample_dates}
\end{figure}

\spara{Non-matching sample generation.}
As for the possible difficulties in distinguishing a non-matching text $\tilde{t}$, we observe that: ($i$) dates $\tilde{t}$ differing from $t$ only for the year are more likely to receive higher similarities if the year in $\tilde{t}$ is present in the training set; and ($ii$) dates with the same digits as $t$ but in different positions tend to be more challenging, since the position plays an important role.
Motivated by these observations, for every matching pair $(I, t)$ (that is originally present in the dataset), we generate a new non-matching pair $(I, \hat{t})$ as follows:
\begin{itemize}
	\item with probability $0.3$, randomly change 1 digit on the \textbf{day};
	\item with probability $0.3$, randomly change 1 digit on the \textbf{month};
	\item with probability $0.15$, randomly change 1 digit out of the last 2 on the \textbf{year};
	\item with probability $0.15$, change the \textbf{year} to a different one, chosen among the years represented in the training set;
	\item with probability $0.1$, pick a random date.
\end{itemize}

\noindent Whenever we change one digit in the day or month, with probability 0.5 we sample the replacement from the set $\{ 0, \ldots , 9 \}$, and with probability 0.5 we sample from the set of digits already present in the date.

\spara{Dataset details.}
We use a real dataset of (matching pairs) of about $50$k images, with dates spanning a period from 2018 to 2021.
We use dates from the year 2022 to experiment on future dates. We perform an \emph{80-10-10} splitting in training, validation and test sets. Sample images are illustrated in Figure~\ref{fig:sample_dates}.
Note that the available ground truth is in the format \emph{dd/mm/yyyy}, while the exact wording inside the cheque might have the year in two digits (\eg, ``21'' instead of ``2021''), and, in case of day or month lower than 10, the zero can be present or not. 
For these reasons, we convert the input text to a normalized format of fixed length without separator (\emph{ddmmyy}), where the possible first zero of day and month is encoded in the  ``~\!$\ast$~\!'' padding character. For instance the date \emph{02/04/2021} is encoded as the text \emph{$\ast$2$\ast$421}. Therefore, the maximum text length is $6$ and the alphabet is \emph{0123456789$\ast$}.
To improve the generalization capabilities of the model in handling future dates, we also enlarge the training dataset with $5$k examples of synthetic dates with year from 2000 to 2030, generated by concatenating digits from MNIST and adding a background resembling the one of cheques, and with $15$k examples of amounts from real cheques with $6$ digits and at most $2$ zeros. 

\spara{Results.}
We report the results of TextMatcher and its competitors on the test set obtained from the splitting of the real dataset of $50$k images (\emph{test-50k}), as well as on a real test set of $165$k examples of real matching and non-matching examples from cheques deposited in January and February 2022 (\emph{real-2022}). 
This way, the models are evaluated not only on future unseen dates, but also on \emph{real negative examples}. The rate of negative examples in this real test set is $1.5\%$, and the handwritten images are $90\%$ of the total.
We assess the performance using the false positive (FP) and false negative (FN) rates.
These metrics are particularly meaningful for the application scenario at hand, in which the main aim is to have low FP while keeping the FN acceptable (\eg $FP \le 2\%$ and $FN \le 20\%$). 
We report the results separately for handwritten ($FP_h$, $FN_h$) and printed  ($FP_p$, $FN_p$) images using the optimal threshold that minimizes: 

\begin{equation}\label{eq:optimal_th}
\small
	10 FP + FN, 
\end{equation}

\noindent with the constraint $FN\le 60\%$ to get reasonable results. 
The results are summarized in Table \ref{table:results_date}.
On the test set with the same distribution as the training set (\emph{test-50k}), TextMatcher and ASTER perform comparably. 
Conversely, in the real test set (\emph{real-2022}), ASTER exhibits a considerable false-negative rate, i.e., $74.66$ and $49.79$ for the handwritten and printed case, respectively, whereas our TextMatcher generalizes well to unseen future dates as well as to the real distribution of non-matching samples, producing error rates comparable to that obtained on \emph{test-50k}. 
These results motivated the adoption of TextMatcher in production for automated cheque processing at UniCredit.

\begin{table}[t]
\vspace{-4mm}
	\centering
	\caption{\small Results on automated cheque processing: matching the issue date of a bank cheque. The optimal threshold $\tau$ (\eq \eqref{eq:threshold}) is selected according to \eq \eqref{eq:optimal_th} achieved by every method on the validation set.}
	\vspace{1mm}
\scriptsize
	\begin{tabular} { p{2.5cm} p{3cm} p{1cm} p{1cm} p{1cm} p{1cm} p{1cm} p{1cm} }
		\hline
		Dataset & Method &  $\tau$ & $FP_p$ & $FN_p$ & $FP_h$ & $FN_h$ \\
		\hline
		\multirow{3}{*}{\emph{test-50k}} & \textbf{TextMatcher} & $0.74$ & $0.33$ & $5.96$ & $0.57$ & $8.23$ \\
		 & ASTER & $0.10$ & $0.08$ & $4.93$ & $0.00$ & $6.76$ \\
		 & Na\"iveTextMatcher & $0.18$ & $8.71$ & $56.48$ & $11.38$ & $61.84$ \\
		\hline
		\multirow{3}{*}{\emph{real-2022}} & \textbf{TextMatcher} & $0.74$ & $0.90$ & $\mathbf{3.55}$ & $0.90$ & $\mathbf{9.59}$ \\
		 & ASTER & $0.10$ & $0.90$ & $49.79$ & $0.00$ & $74.66$ \\
		 & Na\"iveTextMatcher & $0.18$ & $8.80$ & $68.33$ & $7.58$ & $77.26$ \\
		\hline
	\end{tabular}
		\vspace{-3mm}
	\label{table:results_date}
\end{table}


\subsection{General Applicability of Text Matching (IAM Dataset)}\label{experiments:iam}
Here we present experiments carried out on the well-known real public IAM handwriting database~\cite{marti2002iam}. 
This set of experiments aims at investigating the applicability of TextMatcher to more general settings, where the goal is to handle distributions of errors arising from different application scenarios. 
Moreover, these experiments also give the opportunity to highlight the differences between TextMatcher and the text-recognition models, giving the idea of the kind of application in which TextMatcher can achieve remarkable results.

\spara{Non-matching sample generation.}
We consider four ways of injecting errors to generate the negative examples:
\begin{itemize}
	\item \textbf{random}: given a vocabulary $V$,
	the text of a non-matching pair is given by a random word of $V$ (e.g., matching text \emph{meeting}, non-matching text \emph{apple});
	\item \textbf{edit\textsubscript{1}}: the non-matching text has \emph{Levenshtein} distance equal to $1$ from the matching text (e.g., matching text \emph{meeting}, non-matching text \emph{meating});
	\item \textbf{edit\textsubscript{12}}: the non-matching text has \emph{Levenshtein} distance equal to $1$ or $2$ (with equal probability) from the matching text;
	\item \textbf{mixed}: the non-matching text is a random word of $V$ with probability $\tfrac{1}{3}$, has Levenshtein distance equal to $1$ with probability $\tfrac{1}{3}$, or Levenshtein distance equal to $2$ with probability $\tfrac{1}{3}$.
\end{itemize}
We made $4$ datasets containing one non-matching sample for each matching pair.

\spara{Dataset details.}
The IAM handwriting database \cite{marti2002iam} consists of 1\,539 pages of scanned text from 657 different writers. The database also provides the isolated and labeled words that were extracted from the pages of scanned text using an automatic segmentation scheme (and a-posteriori manually verified). We use the dataset at word level, and consider the available splitting proposed for the \emph{Large Writer Independent Text Line Recognition Task}, in which each writer contributed to one set only. 
We set the alphabet to \emph{abcdefghijklmnopqrstuvwxyz-'$\ast$}, and we filter out words with characters outside the alphabet, or words only composed of punctuation marks. Finally, we only retain words with at least $5$ characters.
The final training, validation and test sets have size $17\,550$, $4\,947$, and $4\,175$, respectively.
The maximum word length is $21$. 


\spara{Results.}
We evaluate the selected models using the confusion matrix and the F1-score.
We choose the optimal threshold $\tau$ (\eq \eqref{eq:threshold}) for every method on the validation set, according to the F1-score, and report the performance on the test set. 
The results presented in Table~\ref{table:results} show that the proposed TextMatcher \emph{outperforms all the competitors in all the configurations}.
ASTER was recognized as the best competitor, as expected.
In general, for any method, the best performance is achieved on the $random$ configuration, whereas $edit_{1}$ and $edit_{12}$ configurations have the lowest performance, and $mixed$ has intermediate performance.
This is expected as the higher the similarity between a non-matching text and the corresponding matching text, the higher the difficulty for a model to accomplish the text-matching task, and, hence, the lower the accuracy.

\begin{table}[t]
	\centering
	\vspace{-4mm}
	\caption{\small Results on the IAM handwriting database, for different configurations of non-matching-sample generation. The optimal threshold $\tau$ (\eq \eqref{eq:threshold}) is selected according to the best F1-score achieved by every method on the validation set.}
	\vspace{1mm}
	\scriptsize
	\begin{tabular} { p{2cm} p{3cm} p{1cm} p{1cm} p{1cm} p{1cm} p{1cm} p{1cm} }
		\hline
		Configuration & Method &  $\tau$ & TP & FP & TN & FN & F1 \\
		\hline
		\multirow{3}{*}{\textbf{$random$}} & \textbf{TextMatcher} & $0.46$ & $99.21$ & $1.10$ & $98.90$ & $0.79$ & $\mathbf{99.06}$ \\
		 & ASTER & $0.47$ & $97.39$ & $1.05$ & $98.95$ & $2.61$ & $98.15$ \\
		 & Na\"iveTextMatcher & $0.48$ & $90.28$ & $13.49$ & $86.51$ & $9.72$ & $88.61$ \\
		\hline
		\multirow{3}{*}{\textbf{$edit_1$}} & \textbf{TextMatcher} & $0.48$ & $88.91$ & $18.42$ & $81.58$ & $11.09$ & $\mathbf{85.77}$ \\
		 & ASTER & $0.94$ & $73.63$ & $0.00$ & $100.00$ & $26.37$ & $84.81$ \\
		 & Na\"iveTextMatcher & $0.48$ & $90.06$ & $71.02$ & $28.98$ & $9.94$ & $68.99$ \\
		\hline
		\multirow{3}{*}{\textbf{$edit_{12}$}} & \textbf{TextMatcher} & $0.50$ & $89.84$ & $14.23$ & $85.77$ & $10.16$ & $\mathbf{88.05}$ \\
		& ASTER & $0.94$ & $73.63$ & $0.07$ & $99.93$ & $26.37$ & $84.78$ \\
		& Na\"iveTextMatcher &  $0.46$ & $95.66$ & $75.04$ & $24.96$ & $4.34$ & $70.67$ \\
		\hline
		\multirow{3}{*}{\textbf{$mixed$}} & \textbf{TextMatcher} & $0.52$ & $92.93$ & $8.07$ & $91.93$ & $7.07$ & $\mathbf{92.47}$ \\
		& ASTER & $0.95$ & $73.60$ & $0.05$ & $99.95$ & $26.40$ & $84.77$ \\
		& Na\"iveTextMatcher &  $0.48$ & $82.47$ & $35.78$ & $64.22$ & $17.53$ & $75.57$ \\
		\hline
	\end{tabular}
		\vspace{-3mm}
	\label{table:results}
\end{table}

\subsection{Discussion}\label{sec:advantages}

We conclude this section by highlighting some key advantages of our TextMatcher over its most effective competitor, i.e., the text-recog\-ni\-tion ASTER model. 

The results on IAM show that, in general, TextMatcher achieves better F1-scores than ASTER, in particular for the $edit_{12}$ and $mixed$ configurations, where errors of different complexity need to be recognized together. 
This is related to the distribution of similarities produced by the two models. Indeed, TextMatcher yields a continuous distribution of values, treating different kinds of error similarly. Conversely, ASTER's distribution of similarities is discontinuous, thereby needing different optimal thresholds for different kinds of error. 


Moreover, our TextMatcher can be trained on a specific distribution of errors, and even on different distributions at inference time (through the non-matching texts), as for the case of the issue date of bank cheques.
This means that the model is flexible yet general enough to handle \emph{any particular application scenario}. Indeed, it can be trained on specific patterns that are known to occur and are perhaps particularly difficult to detect for the application scenario at hand (e.g., distinguish ``\emph{facebook ltd}'' from ``\emph{facebook inc}'').
Once trained on the desired negative examples, TextMatcher specializes itself in recognizing these errors, paying more attention to the part of the text that is more relevant for those errors.
Conversely, a text-recognition model like ASTER treats all kinds of error in the same way, thus resulting to be less general and versatile. 

Finally, our TextMatcher is also more efficient than ASTER at inference time.  
We tested the CPU inference time of the two trained models for 1\,000 random examples taken from the IAM \emph{mixed} configuration: ASTER takes around $0.58$ seconds per image on average, whereas TextMatcher takes around $0.07$ seconds per image, which corresponds to a $8.75$x speed-up.

\section{Conclusions}
\label{sec:conc}


In this paper, we study the task of \emph{text matching}, to assess whether an image containing a single-line text corresponds to a given text transcription, and devise the TextMatcher machine-learning model for this task.
The proposed TextMatcher projects image and text into separate embedding spaces, employs a cross-attention mechanism to discover local alignments between those embeddings, and is trained end-to-end on the distribution of errors to be recognized.

We experimentally evaluate TextMatcher on real data, including a proprietary dataset from a real industrial scenario of automated cheque processing, and the popular public IAM dataset.
Compared to a na\"ive model and a state-of-the-art method for the related task of text recognition, TextMatcher proves to be more effective and better suited for handling different distributions of errors.

\bibliographystyle{splncs04}
\bibliography{papers}

\end{document}